\documentclass[]{opendatalab}

\usepackage[utf8]{inputenc}
\usepackage[T1]{fontenc}
\usepackage{microtype}
\usepackage{xspace}

\usepackage[table]{xcolor} 
\usepackage{tcolorbox}
\tcbuselibrary{listings, breakable, skins} 
\usepackage{listings}

\usepackage{colortbl}
\usepackage{longtable}
\usepackage{booktabs}   
\usepackage{multirow}   
\usepackage{graphicx}   
\usepackage{caption}    
\usepackage{wrapfig}

\usepackage{amsmath, amssymb, amsfonts} 
\usepackage{algorithm}     
\usepackage{algpseudocode} 
\usepackage{amssymb}
\usepackage[misc]{ifsym}

\usepackage[numbers]{natbib}
\usepackage{url}
\usepackage{hyperref}
\usepackage[nameinlink]{cleveref}

\definecolor{promptbackground}{RGB}{235, 245, 255}
\definecolor{promptframe}{RGB}{60, 120, 180}
\definecolor{outputbackground}{gray}{0.95}
\definecolor{outputframe}{gray}{0.65}

\newtcblisting{promptbox}[2][]{
    listing only, breakable, title=#2,
    colback=promptbackground, colframe=promptframe, coltitle=white,
    fonttitle=\bfseries, boxrule=0.5pt, arc=3mm, colbacktitle=promptframe,
    fontupper=\small\ttfamily,
    #1
}

\newcommand{\methodname}{AgenticOCR} 

\title{\methodname{}: Parsing Only What You Need for Efficient Retrieval-Augmented Generation}

\author[1,2*]{Zhengren Wang}
\author[1,2*]{Dongsheng Ma}
\author[3*]{Huaping Zhong}
\author[1]{Jiayu Li}
\author[1,2]{\\Wentao Zhang}
\author[1\ \textrm{\Letter}]{Bin Wang}
\author[1\ \textrm{\Letter}]{Conghui He}

\affiliation[1]{Shanghai Artificial Intelligence Laboratory}
\affiliation[2]{Peking University}
\affiliation[3]{SenseTime}

\abstract{
The expansion of retrieval-augmented generation (RAG) into multimodal domains has intensified the challenge for processing complex visual documents, such as financial reports. While page-level chunking and retrieval is a natural starting point, it creates a critical bottleneck: delivering entire pages to the generator introduces excessive extraneous context. This not only overloads the generator's attention mechanism but also dilutes the most salient evidence. Moreover, compressing these information-rich pages into a limited visual token budget further increases the risk of hallucinations.
To address this, we introduce AgenticOCR, a dynamic parsing paradigm that transforms optical character recognition (OCR) from a static, full-text process into a query-driven, on-demand extraction system. By autonomously analyzing document layout in a "thinking with images" manner, AgenticOCR identifies and selectively recognizes regions of interest. This approach performs on-demand decompression of visual tokens precisely where needed, effectively decoupling retrieval granularity from rigid page-level chunking.
AgenticOCR has the potential to serve as the "third building block" of the visual document RAG stack, operating alongside and enhancing standard Embedding and Reranking modules. Experimental results demonstrate that AgenticOCR improves both the efficiency and accuracy of visual RAG systems, achieving expert-level performance in long document understanding. Code and models are available at \url{https://github.com/OpenDataLab/AgenticOCR}.
}

\metadata[* Equal contribution\quad $\textrm{\Letter}$ Corresponding author]{}
\correspondence{Conghui He, \email{heconghui@pjlab.org.cn}}
\date{\today}

\begin{document}

\maketitle

\section{Introduction}
Optical character recognition (OCR) has evolved into a remarkably mature field. Driven by advances in vision-language models (VLMs), billion-parameter models such as MinerU2.5 (1.2B) and PaddleOCR-VL-1.5 (0.9B) routinely achieve 90–95\% accuracy on OmniDocBench~\citep{ouyang2025omnidocbench}, while even ultra-lightweight variants like OpenOCR (0.1B) reach the 90\% threshold. These significant breakthroughs signify that---traditional, full-document OCR or parsing task is largely solved.

Yet this very success has created a paradox: while the supply of general-purpose, full-document parsing capability nears saturation, a critical demand gap for instruction-following or on-demand parsing remains unmet in downstream applications, especially in retrieval‑augmented generation (RAG) over visual documents such as financial reports, technical manuals, and scholarly articles.

In visual RAG pipelines, documents are typically chunked and retrieved at the page level—a natural but coarse granularity \citep{wang2025vrag, yuvisrag}. It delivers entire document pages—often containing headers, footers, decorative elements, and irrelevant sections—to the generator. This mismatch between information need and retrieval granularity introduces two compounding inefficiencies. First, the generator's attention mechanism becomes diluted by extraneous visual context, reducing its capacity to focus on query-relevant evidence. Second, and more critically, high-resolution document pages must be compressed into limited visual token budgets, inevitably sacrificing fine-grained details and amplifying hallucination risks—particularly for rotated tables, small-font annotations, intricate formulas or complex layouts. Furthermore, recent studies on multi-image reasoning highlight that VLMs struggle with evidence localization in visual RAG and are easily misled by irrelevant visual information \citep{sun2025visrag, wang2025vrag}, underscoring the inherent limitations of page-level retrieval.

\begin{figure}[tb]
\centering
\includegraphics[width=\textwidth]{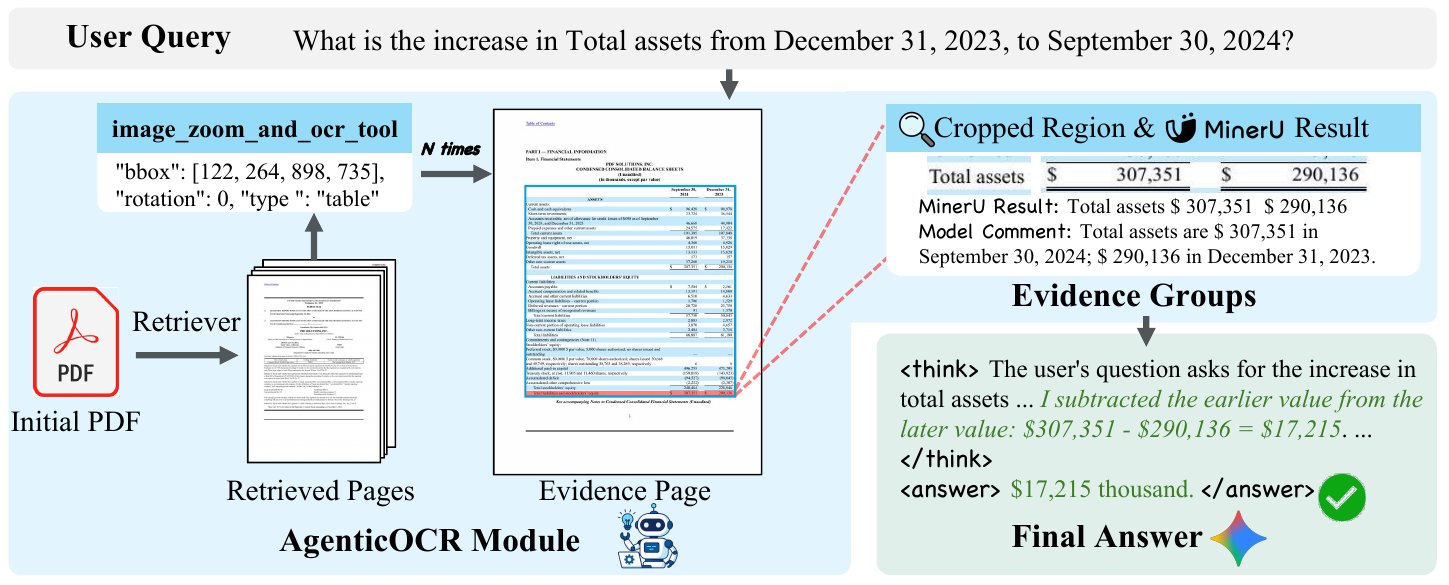}
\caption{Overview of AgenticOCR-based RAG. AgenticOCR performs on-demand decompression of visual information precisely where it is needed by utilizing operations such as zoom and rotate.}
\label{fig:workflow}
\end{figure}

This challenge cannot be resolved by further improving full-page OCR accuracy alone, but calls for a shift from "parsing everything" (base model) to "parsing only what you need" (instruction-following model). Here, we introduce and formalize this paradigm as AgenticOCR: a dynamic, query-driven framework that elevates OCR from a static pre-processing step into an intelligent, agentic process. By autonomously reasoning over document layout and content, an AgenticOCR model identifies, locates, and selectively parses the regions of interest relevant to a specific query or instruction. As shown in Fig. \ref{fig:workflow}, with zoom, rotate and other operations, it performs on-demand decompression of visual information precisely where needed. Meanwhile, acting as a powerful filter, it can serve as a computationally efficient pre-processor for large generative models like Gemini and GPT, reducing their visual token consumption and improving overall system efficiency.

\paragraph{Contributions} Our main contributions are threefold:
\begin{itemize}
    \item \textbf{Conceptual Formalization}: We pioneer to introduce AgenticOCR, positioning it as a potential "third building block" in the visual document RAG stack—alongside Embedding and Reranking modules. Beyond RAG, AgenticOCR also holds promise for key information extraction (KIE), element-level evidence citation, and interactive assistants.
    \item \textbf{Model Realization}: We implement high-performing AgenticOCR models through a two-stage training. Based on Qwen3-VL and a specialized "image\_zoom\_and\_ocr\_tool", we first use rejection sampling to distill high-quality trajectories from Gemini-3-Pro-Preview as a cold start. Subsequently, we apply Group Relative Policy Optimization (GRPO) with specialized reward functions for alignment. Both the constructed datasets and models are open-sourced.
    \item \textbf{Empirical Validation}: Our experiments demonstrate the effectiveness of the AgenticOCR module. It improves the signal-to-token ratio in visual RAG pipelines, leading to significant accuracy gains. We also discuss future optimization directions to inspire further research.
\end{itemize}

\section{Related Work} 
\label{sec:RelatedWork}
\paragraph{Visual Retrieval-Augmented Generation}
The paradigm of Retrieval-Augmented Generation (RAG), initially successful in textual domains, has been extended to handle visually rich documents to mitigate model hallucination and knowledge limitations \citep{asai2024self, lewis2020retrieval}. Early visual RAG systems, such as ColPali \citep{faysse2024colpali} and VisRAG \citep{yu2024visrag}, pioneered the use of document page snapshots as retrieval units, directly feeding images to VLMs to bypass error-prone OCR pipelines and preserve crucial layout semantics. However, this page-level granularity creates a fundamental bottleneck: retrieving entire pages introduces substantial irrelevant visual content, which dilutes the generator’s attention and forces high-resolution pages into limited visual token budgets, often sacrificing detail and increasing hallucination risk \citep{tanaka2025vdocrag, wang2025vidorag}. Recent efforts aim to refine this granularity. RegionRAG~\citep{li2025regionrag} shifts retrieval from the document to the region level, using weakly-supervised contrastive learning and post-processing to identify and group relevant patches into regions. For long-document understanding, frameworks like DocLens \citep{zhu2025doclens} employ multi-agent systems to first navigate to relevant pages and then identify specific visual elements (tables, charts), performing hierarchical evidence localization with Gemini models. Similarly, DocDancer~\citep{zhang2026docdancer} formulates DocVQA as an agentic information-seeking problem, utilizing search and read tools for iterative document exploration and localized comprehension. Notably, VRAG-RL~\citep{wang2025vrag} and EVisRAG~\citep{sun2025visrag} integrate reinforcement learning to teach models to interleave retrieval, visual perception, and reasoning, aiming for more faithful and efficient use of multi-image evidence. Despite these advances, there is no specialized or lightweight model for on-demand evidence extraction, making parsing itself a query-driven process.

\paragraph{OCR and Document Parsing} 
Optical Character Recognition (OCR) and document parsing have reached remarkable maturity. Traditional pipeline approaches decompose parsing into sequential modules, offering interpretability at the cost of error propagation and operational complexity \citep{zhang2024document,wang2024mineru}. The rise of vision-language models has spurred end-to-end architectures \citep{blecher2023nougat,MinerU2,wei2024general,dots.ocr} that unify parsing within a single model, eliminating cascaded failures but introducing new challenges: native-resolution processing incurs $\mathcal{O}(N^2)$ computational complexity, while token redundancy from blank regions wastes precious visual token budgets. Recent multi-stage methods \citep{feng2025dolphin, li2025monkeyocr, zhang2025monkeyocr, cui2025paddleocrvl} enhance efficiency and accuracy by decoupling layout analysis from content recognition—first analyzing downsampled pages globally, then parsing cropped regions locally. MinerU2.5 \citep{niu2025mineru2} exemplifies this direction with its coarse-to-fine inference mechanism, while PaddleOCR-VL-1.5 \citep{cui2026paddleocrvl15multitask09bvlm} and HunyuanOCR \citep{team2025hunyuanocr} enhance robustness against physical distortions in the wild settings. Meanwhile, DeepSeek-OCR \citep{wei2025deepseek} pioneers a complementary perspective: treating visual tokens as a compression medium for textual information, achieving 7–20$\times$ token reduction while preserving decoding fidelity. Its successor, DeepSeek-OCR 2 \citep{wei2026deepseek2}, further advances this paradigm by modeling visual causal flow through LLM-based encoders. Despite remarkable progress in full-document parsing accuracy (90–95\% on OmniDocBench), these systems remain fundamentally static. This rigidity becomes particularly problematic in RAG contexts where only sparse regions are query-relevant, motivating our shift toward dynamic, instruction-following parsing.

\paragraph{Agentic Multimodal Models}
The evolution of multimodal reasoning has witnessed a paradigm shift from passive perception to active agency, where models dynamically interact with visual inputs through tool invocation rather than merely interpreting static representations. Pioneering work by OpenAI's o3~\citep{su2025thinking} introduced the concept of ``thinking with images,'' demonstrating that interleaving visual manipulation with textual reasoning enables more faithful visual cognition. Subsequent research has sought to replicate and extend this capability through reinforcement learning. DeepEyes~\citep{zheng2025deepeyes} incentivizes region-focused visual exploration via an image zoom-in tool, allowing models to autonomously identify and crop regions of interest during reasoning without cold-start supervision. Thyme~\citep{zhang2025thyme}, PyVision~\citep{zhao2025pyvision} and CodeVision~\citep{guo2025thinking} further expand functionality by enabling code-based image operations (e.g., rotation, contrast adjustment, and multi-step transformations), achieving rich visual interaction through end-to-end SFT+RL training. Meanwhile, DeepEyesV2~\citep{hong2025deepeyesv2} explores integrating external knowledge acquisition via online search, recognizing that perception alone is insufficient for knowledge-intensive tasks. Despite these advances, existing agentic multimodal frameworks remain predominantly oriented toward general visual reasoning—such as object recognition and spatial inference—without addressing the parsing demands of document intelligence. Crucially, they treat visual tools as reasoning aids rather than extraction mechanisms. AgenticOCR bridges this gap by specializing the agentic paradigm for document parsing.

\section{AgenticOCR: Parsing Only What You Need}

\label{sec:method}
We present \methodname{}, a dynamic parsing paradigm that transforms OCR from a static pre-processing step into a query-driven, agentic process. Our approach centers on three core components: (1) a unified visual interaction tool that enables fine-grained, on-demand evidence extraction; (2) a SFT+RL training pipeline that first establishes a robust tool-use policy and then optimizes precise region localization and preference alignment; (3) An integration protocol that allows \methodname{} to serve as a plug-and-play module within visual RAG pipelines. Fig.~\ref{fig:workflow} illustrates the overall workflow.

\subsection{The \texttt{image\_zoom\_and\_ocr\_tool} Primitive}
\label{subsec:tool}

For flexible parsing, we design a single but versatile tool that unifies region localization, geometric correction, and content recognition into an atomic operation. Formally, the tool function is defined as:

\begin{equation}
\mathcal{T}: (I, \mathbf{b}, \theta, \tau) \mapsto (I_{\text{crop}}, \mathcal{R}),
\end{equation}

where $I$ is the input page image; $\mathbf{b} = [x_{\min}, y_{\min}, x_{\max}, y_{\max}]$ denotes a bounding box in normalized coordinates (0--1000) relative to the original page image, specifying the region to crop; $\theta \in \{0^\circ, 90^\circ, 180^\circ, 270^\circ\}$ specifies rotation angle applied counter-clockwise to the cropped region; and $\tau \in \{\texttt{region}, \texttt{text}, \texttt{table}, \texttt{image}, \texttt{equation}\}$ indicates the semantic type of the cropped images. The tool returns both the cropped image $I_{\text{crop}}$ and structured recognition results $\mathcal{R}$. Crucially, the tool's behavior depends on the specified type $\tau$:

\begin{itemize}
    \item \textbf{Region mode ($\tau=region$):} Invokes MinerU's full two-stage pipeline—first performing layout analysis on the zoomed-in region, then executing fine-grained recognition on detected sub-elements. This mode is designed to \textbf{enhance active perception for complex regions}, splitting regions of interest from the whole page. Notably, the tool returns layout results with coordinates normalized to the original page rather than the cropped image. This circumvents confusion between different coordinate systems and facilitate the tracing of evidence.

    \item \textbf{Element mode ($\tau \in \{text, table, equation\}$):} Bypasses layout detection and directly applies MinerU's recognition module to the cropped region. This enables the model to directly parse target elements when layout is clear, thereby improving efficiency and avoiding potential errors arising from unnecessary layout detection steps.
    \item \textbf{Image mode ($\tau=image$):} Returns only the cropped visual patch without OCR, enabling pure visual perception before committing to content extraction.
\end{itemize}

This design embodies a fundamental shift: OCR becomes an active perception process rather than a passive preprocessing step. The model decides \textit{where to look}, \textit{how to orient}, and \textit{at what semantic granularity to parse}—mimicking human visual attention during document reading.

\begin{figure*}[tb]
\centering
\includegraphics[width=\textwidth]{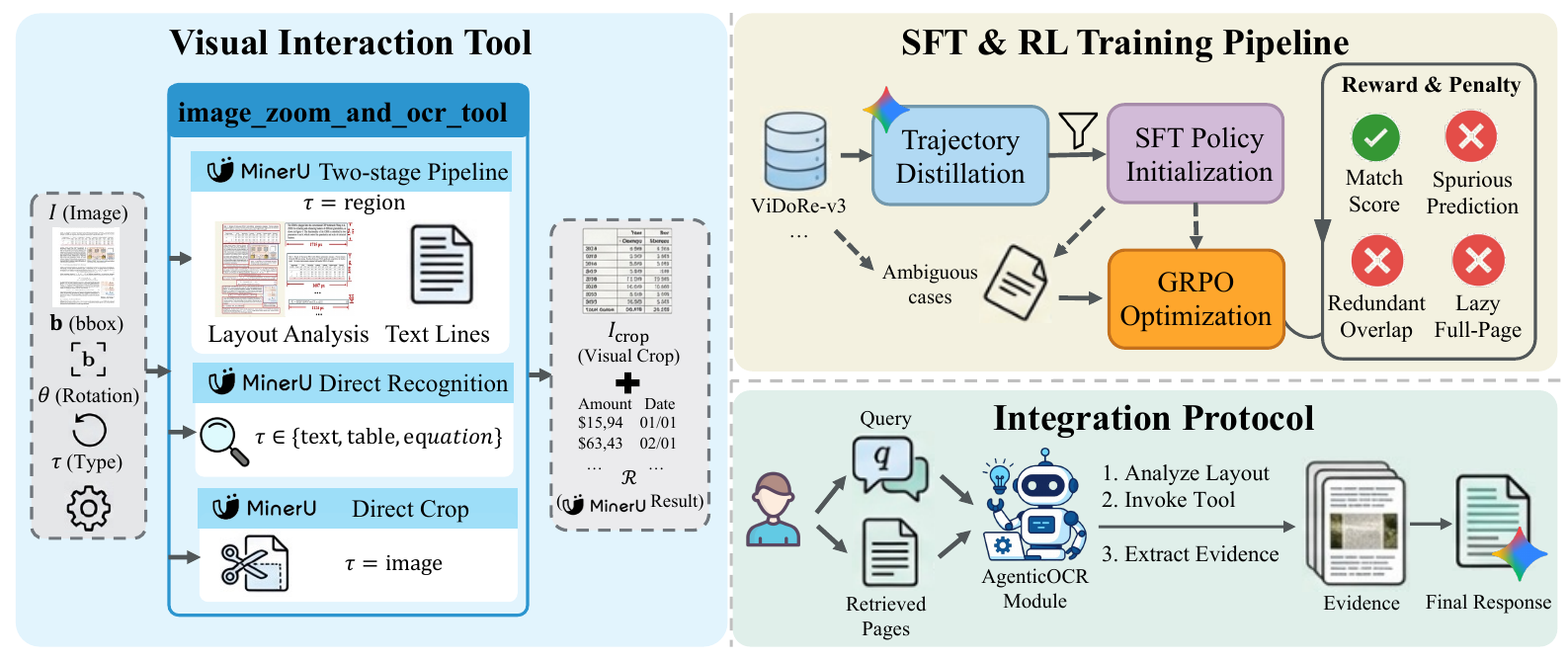}
\caption{The training and inference of AgenticOCR. Built upon the \texttt{image\_zoom\_and\_ocr\_tool}, trajectory distillation is first performed to initialize the SFT policy. The model is subsequently optimized through GRPO and finally deployed through an integration protocol within visual RAG pipelines.}
\label{fig:method}
\end{figure*}

\subsection{Cold-Start via Supervised Fine-Tuning}
\label{subsec:sft}

Direct RL training often fails due to sparse rewards and unstable exploration~\citep{hong2025deepeyesv2}. We therefore employ a cold-start phase using high-quality trajectories distilled from Gemini via rejection sampling~\citep{wang2025rare,zhang2026docdancer}. Our goal is not maximal performance but to instill a stable prior for when and how to invoke the tool.

\paragraph{Trajectory Distillation via Rejection Sampling.} 
We construct SFT trajectories from the ViDoRe-v3 benchmark~\citep{loison2026vidore}, which contains queries, document pages, and human-annotated bounding boxes  $\mathcal{B}^{\text{gt}}$ for relevant regions. To obtain high-quality, multi-turn reasoning traces with effective tool invocations, we employ rejection sampling on Gemini-3-Pro-Preview \citep{team2023gemini}. However, raw annotations exhibit inconsistent granularity (from full-page to single-cell level), making them unsuitable for direct application. Instead, we employ a dual-threshold strategy to filter high-quality trajectories:

Given ground-truth boxes \(\mathcal{B}^{\text{gt}}\) and predicted boxes \(\mathcal{B}^{\text{pred}}\), we define two complementary recall metrics:
\begin{equation}
\mathrm{IoU}_{\mathrm{EM}}(b^{\text{gt}}, b^{\text{pred}}) = \frac{|b^{\text{gt}} \cap b^{\text{pred}}|}{|b^{\text{gt}} \cup b^{\text{pred}}|}, \quad \mathrm{IoU}_{\min}(b^{\text{gt}}, b^{\text{pred}}) = \frac{|b^{\text{gt}} \cap b^{\text{pred}}|}{\min(|b^{\text{gt}}|, |b^{\text{pred}}|)}, \\
\end{equation}
\begin{align}
\mathrm{Recall}_{\mathrm{EM}} &= \frac{1}{|\mathcal{B}^{\text{gt}}|} \sum_{b^{\text{gt}} \in \mathcal{B}^{\text{gt}}} \mathbb{I}\Big( \max_{b^{\text{pred}} \in \mathcal{B}^{\text{pred}}} \mathrm{IoU}_{\mathrm{EM}}(b^{\text{gt}}, b^{\text{pred}}) \geq thres_{\mathrm{EM}} \Big), \\
\mathrm{Recall}_{\min} &= \frac{1}{|\mathcal{B}^{\text{gt}}|} \sum_{b^{\text{gt}} \in \mathcal{B}^{\text{gt}}} \mathbb{I}\Big( \max_{b^{\text{pred}} \in \mathcal{B}^{\text{pred}}} \mathrm{IoU}_{\min}(b^{\text{gt}}, b^{\text{pred}}) \geq thres_{\mathrm{\min}} \Big),
\end{align}
where \(\mathrm{IoU}_{\min}\) uses the minimum box area as denominator (tolerant to granularity mismatch) and \(\mathrm{IoU}_{\mathrm{EM}}\) is the standard intersection-over-union (sensitive to precise alignment). We adopt $thres_{\mathrm{EM}} = 0.6$ and  $thres_{\mathrm{\min}} = 0.8$ based on our experience. We first filter trajectories with \(\mathrm{Recall}_{\min} < 0.8\) to ensure coverage of relevant regions, then rank survivors by \(\mathrm{Score} = \mathrm{Recall}_{\min} + \mathrm{Recall}_{\mathrm{EM}}\) to prioritize spatial precision. This yields 5k high-quality positive trajectories with diverse tool usage patterns.

\paragraph{Negative Sampling for Page-Level Filtering.}
To suppress spurious tool calls and evidence returns on irrelevant pages, we construct 1.5K negative ViDoRe-v3 samples by collecting semantically similar but not logically related pages. For each query, we remove its ground-truth page from the source PDF, rerank remaining pages using Qwen3-VL-Reranker-8B~\citep{li2026qwen3}, and select hard negative candidates with relevance scores in \([0.05, 0.30]\). We further filter out false negatives via Gemini verification, 
\begin{wrapfigure}{r}{0.4\textwidth}
\vspace{-1em}
    \centering
    \includegraphics[width=\linewidth]{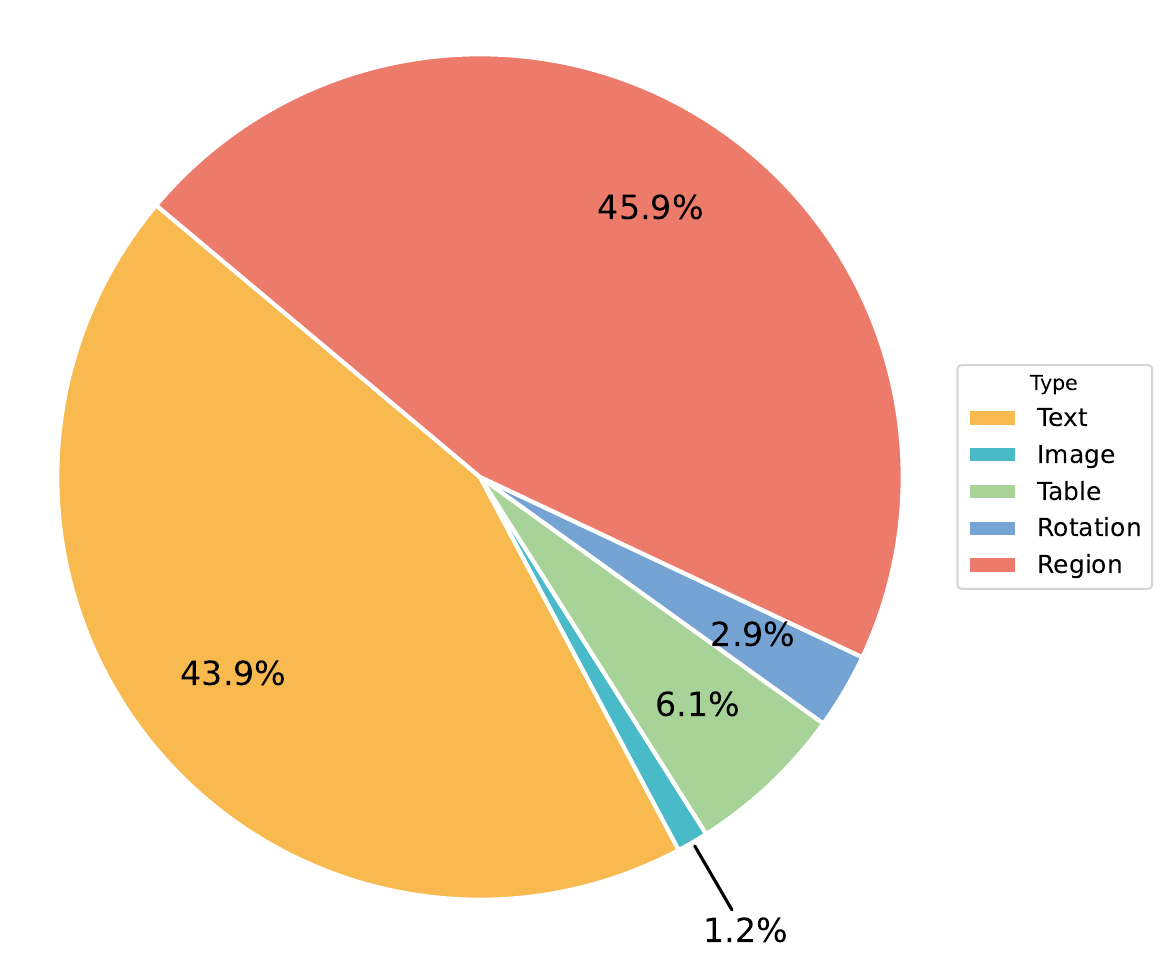}
    \caption{Distribution of positive trajectories in SFT dataset.}
    \label{fig:tool_type_pie}
\vspace{-2em}
\end{wrapfigure}
retaining 1.5k pairs that contain no query-relevant evidence, whose distribution is consistent with the pages returned by the reranker during inference. These negatives teach the model to suppress evidence returns on irrelevant pages—a critical capability for RAG integration. The final SFT dataset holds a 77:23 positive-to-negative ratio to balance recall preservation with precision.

\paragraph{Training Protocol.}
We train on interleaved multi-turn trajectories with token-wise loss masking: only assistant-generated tokens (reasoning steps and tool calls) contribute to the loss; user prompts and tool observation tokens are masked out. This focuses learning on decision-making rather than observation reproduction, aligning with standard practice in agentic training~\citep{zheng2025deepeyes, wang2025code}.

\subsection{Alignment via Reinforcement Learning}
\label{subsec:rl}

While SFT establishes basic tool-use competence, it cannot optimize for spatial precision or human preference alignment. We apply Group Relative Policy Optimization (GRPO)~\citep{shao2024deepseekmath} with a tailored reward function that jointly optimizes coverage, localization accuracy, and behavioral constraints.

\paragraph{Curriculum Data Selection.}
To maximize training efficiency, we curate a curriculum that focuses RL on ambiguous cases where the SFT policy exhibits unstable behavior. We first augment the SFT dataset with 1k samples from CodeVision with rotated/complex layouts~\citep{wang2025code} (partly de-rotated to match real-world distributions) and apply uncertainty-based difficulty filtering: for each candidate, we collect 8 rollouts from the SFT model and retain samples with high standard deviation in \(\mathrm{Score} = (\mathrm{Recall}_{\min} + \mathrm{Recall}_{\mathrm{EM}})/2\). This uncertainty-based filtering concentrates optimization on the most informative examples, avoiding wasted computation on already-converged easy cases or hopelessly difficult outliers. The final RL dataset contains 2.5k positives from ViDoRe-v3, 1k positives from CodeVision, 1k negative samples from ViDoRe-v3 (about 20\% ratio).

\paragraph{Reward Design.} For positive samples, the reward combines dual-recall metrics with penalties:

\begin{equation}
R_{\text{pos}} = \underbrace{\frac{\mathrm{Recall}_{\min} + \mathrm{Recall}_{\text{EM}}}{2}}_{\text{accuracy}} - \underbrace{P_{\text{over-pred}} - P_{\text{overlap}} - P_{\text{oversized}}}_{\text{behavioral constraints}}.
\end{equation}

\begin{enumerate}
    \item \textbf{Spurious Prediction Penalty ($P_{\text{over-pred}}$).} Predictions that fail to overlap meaningfully with any ground-truth region (i.e., low $\max_{b^{\text{gt}}} \mathrm{IoU}_{\min}(b^{\text{pred}}, b^{\text{gt}})$) indicate unfounded hallucinations and low precision. We penalize such spurious boxes with escalating severity:
    \begin{equation}
    P_{\text{over-pred}} = 
    \begin{cases}
        0.05 & \text{if 1 spurious box},\\
        0.20 & \text{if 2 spurious boxes},\\
        0.30 & \text{if $\geq$3 spurious boxes}.
    \end{cases}
    \end{equation}
    
    \item \textbf{Redundant Overlap Penalty ($P_{\text{overlap}}$).} Overlapping predictions (high $\mathrm{IoU}_{\mathrm{EM}}$ between any pair) waste computational resources and fragment evidence representation. We apply the same escalating penalty structure as $P_{\text{over-pred}}$ to redundant boxes, encouraging the model to consolidate evidence into minimal, non-redundant regions.
    
    \item \textbf{Lazy Full-Page Parsing Penalty ($P_{\text{oversized}}$).} When invoking the \texttt{region} type, predictions covering $>85\%$ of the page area degenerate to full-page parsing—defeating the purpose of query-driven extraction. We impose a fixed penalty of $0.10$ for such oversized regions, incentivizing the model to develop genuine spatial reasoning rather than relying on the underlying layout detector's coarse output.
\end{enumerate}

These penalties collectively enforce a principled perception: extract the minimal set of regions necessary to answer the query, with high spatial fidelity and no redundancy. For negative samples, we simply adopt a binary reward: \(R_{\text{neg}} = 1\) if no boxes predicted, else \(0\). This sharp signal reinforces page-level relevance judgement.

\subsection{Integration with Visual RAG Pipelines}
\label{subsec:integration}

\methodname{} operates as a plug-and-play middleware between page retrieval and generation modules (Figure~\ref{fig:workflow}). Given a query $q$ and top-$k$ retrieved pages $\{p_i\}_{i=1}^k$ from visual page retrievers (e.g., Qwen3-VL-Reranker or Qwen3-VL-Embedding~\citep{li2026qwen3}), \methodname{} processes each page independently:

\begin{enumerate}
    \item The agent analyzes the page layout and query semantics to decide whether parsing is needed.
    \item If needed, it invokes \texttt{image\_zoom\_and\_ocr\_tool} one or more times to extract structured evidence $\mathcal{E}_i = \{(b_j, \tau_j, \text{OCR}_j)\}_j$.
    \item Extracted evidence is formatted as interleaved text–image snippets and passed to the generator such as Gemini, optionally with resolution-reduced page screenshots.
\end{enumerate}

This design achieves two critical advantages: (1) it preserves visual context for layout-sensitive reasoning while eliminating token waste on irrelevant regions; and (2) it requires no modification to existing strong generators—only a minor prompt engineering to accept structured evidence inputs. As shown in Section~\ref{sec:experiments}, this simple integration yields substantial gains in accuracy.

\section{Experiments}
\label{sec:experiments}

\subsection{Experimental Setup}
\label{sec:exp_setup}

\paragraph{Benchmarks and Metrics.}
We evaluate our visual RAG system on two challenging benchmarks: MMLongBench-Doc~\citep{ma2024mmlongbench} and FinRAGBench-V~\citep{zhao2025finragbench}. MMLongBench-Doc tests reasoning over lengthy, multi-domain documents (avg. 49.4 pages) that require integrating scattered information across diverse modalities. Its ``Unanswerable'' subset directly evaluates hallucination response. Performance of human experts on this benchmark is reported as 65.8. FinRAGBench-V is helpful for our analysis due to two unique features: its use of documents with dense, newspaper-like layouts, and its support for evaluating visual citation (pinpointing block-level evidence). We adhere to the original evaluation protocols: rule-based scoring for MMLongBench-Doc and an LLM-as-a-judge for FinRAGBench-V. Further statistics and evaluation details are provided in Appendices~\ref{sec:app_data}, \ref{sec:app_mmlong_eval}, and~\ref{sec:app_finrag_eval}.

\paragraph{Models and Baselines.}
We train two variants of our proposed modules: \methodname-4B and \methodname-8B, initialized from Qwen3-VL-4B-Instruct and Qwen3-VL-8B-Instruct respectively. The visual RAG performance is benchmarked against three categories of baselines, whose scores are excerpted from DocLens~\citep{zhu2025doclens}. (1) \textbf{Vanilla VLMs}: Models that receive only page screenshots without any additional processing. This includes GPT-4o, Claude-4-Sonnet, Gemini-2.5-Flash, Gemini-2.5-Pro, and o4-mini; (2) \textbf{VLMs augmented with OCR}: The same VLMs are provided with both page screenshots and OCR-extracted text appended to each page; (3) \textbf{VLM-based Agentic Frameworks}: We compare with MACT~\citep{yu2025visual}, M3DocRAG~\citep{cho2024m3docrag}, MDocAgent~\citep{han2025mdocagent}, SimpleDoc~\citep{jain2025simpledoc}, and DocLens~\citep{zhu2025doclens}. Performance numbers for MACT, M3DocRAG, and MDocAgent are taken from best results in their original papers.

\begin{wrapfigure}{r}{0.4\textwidth}
\vspace{-1em}
\centering
\includegraphics[width=\linewidth]{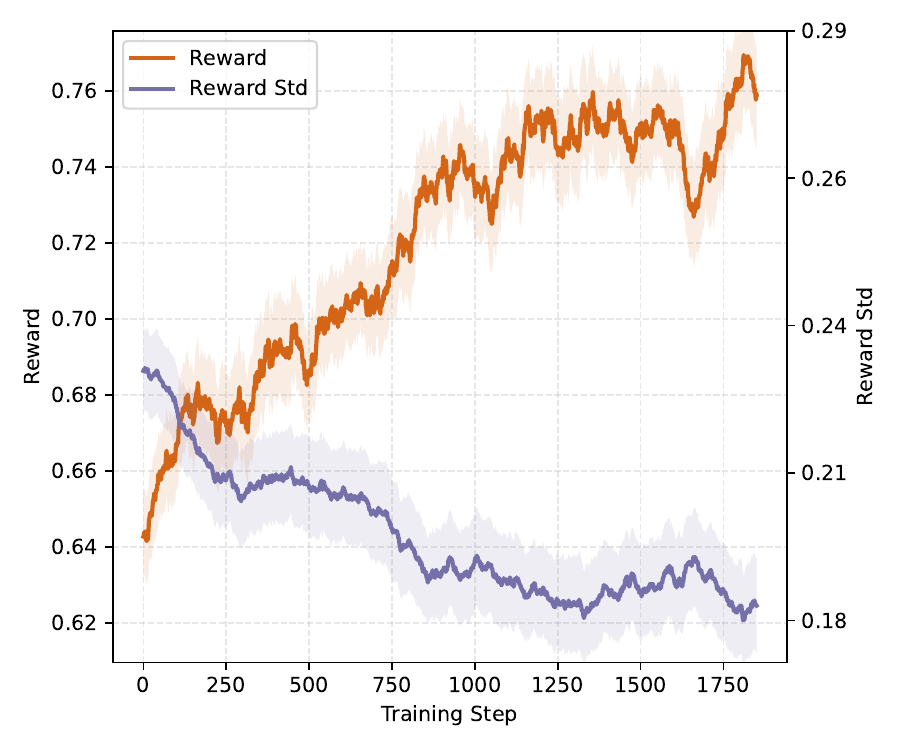}
\caption{RL curves for reward and its standard deviation. The trend demonstrates that the agent effectively learns to use tools strategically to solve tasks.}
\vspace{-1em}
\label{fig:loss_curve}
\end{wrapfigure}
\paragraph{Implementation Details.}
For model training, we train \methodname-4B and \methodname-8B in two stages: supervised fine-tuning (SFT) followed by reinforcement learning (RL) with GRPO. For SFT, we use a learning rate of \(1 \times 10^{-5}\) and train for 6 epochs. The total batch size is 64 with gradient accumulation steps of 8. Early stopping is applied when the validation loss plateaus. Starting from the SFT checkpoint, we perform RL for 2 epochs with a learning rate of \(1 \times 10^{-6}\) and a total batch size of 64. For each query, we sample 16 rollouts. The KL divergence coefficient is set to 0.01 to constrain policy updates and maintain training stability. The reward follows the design in Section~\ref{subsec:rl}. 

For inference, we use all page images from the corresponding PDF as the retrieval pool for each query, employing Qwen3-VL-Reranker-8B for top-$k=30$ page retrieval. Actually, a top-$k$ of 20 already achieves a 96.7\% page recall on MMLongBench-Doc. For MMLongBench-Doc, we adopt an expanded retrieval strategy: for pages in the top-$k$ set, if an adjacent page (immediately preceding or following) is not already included, it is added. These pages are then processed independently by the AgenticOCR module. To improve the recall rate of our current AgenticOCR model on MMLongBench-Doc, we attempt extraction up to three times for each page; a page is only deemed irrelevant if all attempts yield no relevant evidence. 

For the generator (Gemini-2.5-Pro), we evaluate four input configurations to investigate the contribution of each component: (1) Page: only selected page images (resized to \(1024\times1024\) maximum) are provided; (2) Page+OCR: selected page images together with their complete OCR text; (3) Evidence: low-resolution page images plus cropped evidence images (resized to \(512\times512\) maximum) extracted by AgenticOCR, without OCR text; (4) Evidence+OCR: besides page images, both cropped evidence images and their corresponding OCR texts (raw MinerU output and AgenticOCR model-generated comments) are supplied. For more implementation details, please refer to Appendix~\ref{sec:app_implement}.

\subsection{Main Results}
\label{sec:exp_main}

Table~\ref{tab:exp_main_results} presents the main results on both benchmarks. Our proposed \methodname{} achieves competitive performance across the board, with several key observations.

\begin{table*}[tbp]
\centering
\footnotesize
\resizebox{\textwidth}{!}{

\begin{tabular}{@{}l cccccc|c| ccc|c@{}}
\toprule
\multirow{2}{*}{\textbf{Model}} & \multicolumn{7}{c}{\textbf{MMLongBench-Doc}} & \multicolumn{4}{c}{\textbf{FinRAGBench-V}} \\
\cmidrule(r){2-8} \cmidrule(l){9-12}
& \textbf{TXT} & \textbf{LAY} & \textbf{CHA} & \textbf{TAB} & \textbf{FIG} & \textbf{UNA} & \textbf{ALL} & \textbf{TXT} & \textbf{TAB} & \textbf{CHA} & \textbf{ALL} \\ \midrule
\multicolumn{12}{@{}c}{\textit{Vanilla VLMs}} \\ \midrule
GPT-4o$^{\dag}$        & 46.3 & 46.0 & 45.3 & 50.0 & 44.1 & 20.2 & 42.8 & - & - & - & 37.2 \\
Claude-4-Sonnet        & 50.4 & 49.4 & 50.5 & 57.3 & 43.9 & 59.0  & 53.4 & 36.6 & 20.2 & 51.9 & 33.8 \\
Gemini-2.5-Flash & 44.0 & 53.2 & 46.0 & 43.9 & 48.2 & 56.7 & 49.6 & 49.0 & 41.6 & 41.0 & 43.0 \\
Gemini-2.5-Pro & 52.1 & 62.1 & 55.5 & 55.3 & 54.0 & 59.9 & 58.1 & 62.2 & 55.3 & 50.4 & 54.9 \\ 
o4-mini$^{\dag}$       & - & - & - & - & - & - & - & - & - & - & 62.4 \\ \midrule
\multicolumn{12}{@{}c}{\textit{VLMs Augmented with OCR}} \\ \midrule
Claude-4-Sonnet   &  52.7 & 51.6 & 50.0 & 58.1 &  45.3 & 65.9 & 56.0 & 58.7 & 21.6 & 54.3 & 41.0 \\
Gemini-2.5-Flash & 55.9 & 54.9 & 52.7 & 63.4 & 50.3 & 60.8 & 58.5 & 67.6 & 64.4 & 46.1 & 58.3 \\
Gemini-2.5-Pro & 59.7 & 65.3 & 60.8 & \underline{68.3} & 55.7 & 58.4 & 63.3 & 70.0 & 70.0 & 56.2 & 64.9 \\ \midrule
\multicolumn{12}{@{}c}{\textit{VLM-based Agentic Frameworks}} \\ \midrule
MACT (w/ MiMo-VL-7B)$^{\dag}$  & - & - & - & - & - & - & 47.4 & - & - & - & - \\
M3DocRAG (w/ Qwen2-VL-7B)$^{\dag}$  & 30.0 & 23.5 & 18.9 & 20.1 & 20.8 & 5.8 & 21.0  & - & - & - & - \\
MDocAgent (w/ GPT-4o)$^{\dag}$  & - & - & - & - & - & - & 42.0 & - & - & - & - \\
\multicolumn{12}{@{}l}{\textbf{SimpleDoc}} \\
\quad w/ Claude-4-Sonnet & 52.1 & 53.3 & 58.3 & 62.4 & 46.9 & 66.5 & 58.6 &  59.6 & 68.9 & 54.9 & 61.7 \\
\quad w/ Gemini-2.5-Flash & 45.5 & 57.4 & 49.0 & 51.6 & 45.2 & 66.5 & 53.3 & 70.2 & 56.2 & 53.6 &  58.3 \\
\quad w/ Gemini-2.5-Pro & 48.4 & 54.8 & 55.7 & 56.1 & 52.5 & 59.7 & 56.6 & 67.5 & 64.0 & 60.9 & 63.6 \\
\multicolumn{12}{@{}l}{\textbf{DocLens}} \\
\quad w/ Claude-4-Sonnet & 59.9 & 58.2 & 54.4 & 63.9 & 55.3 & \textbf{74.0} & 63.3 & 70.2 & 66.0 & 60.3 & 64.8 \\
\quad w/ Gemini-2.5-Flash & 59.5 & 61.5 & 54.8 & 66.9 & 59.0 & \underline{73.8} & 64.7 & 69.9 & 71.3 & 64.5 & 68.5 \\
\quad w/ Gemini-2.5-Pro &  63.7 & 64.6 & \textbf{64.3} & \textbf{69.7} & 60.2 & 72.2 & \textbf{67.6$^{*}$} & 68.9 & 74.2 & 67.1 & 70.4 \\
\multicolumn{12}{@{}l}{\cellcolor{gray!15}\textbf{\methodname{} (Ours)}} \\
\multicolumn{12}{@{}l}{\quad \textit{Agent Backbone: AgenticOCR-4B}} \\
\quad\quad w/ Page & 57.9 & 64.5 & 59.5 & 57.3 & 58.0 & 63.3 & 61.6 & 85.4 & 68.5 & 65.7 & 71.1 \\
\quad\quad w/ Page + OCR & 62.8 & \underline{67.3} & 58.9 & 64.4 & 57.8 & 66.0 & 64.1 & \underline{89.1} & \textbf{77.4} & 66.8 & 76.0 \\
\quad\quad w/ Evidence & 61.6 & 66.6 & \underline{63.8} & 62.8 & \underline{63.1} & 63.2 & 64.7 & 86.8 & 75.5 & 73.5 & 77.2 \\
\quad\quad w/ Evidence + OCR & 62.4 & 62.3 & 62.5 & 62.3 & 62.3 & 66.6 & 64.9 & 89.0 & 74.8 & 73.4 & 77.4 \\
\multicolumn{12}{@{}l}{\quad \textit{Agent Backbone: AgenticOCR-8B}} \\
\quad\quad w/ Page & 57.5 & 60.7 & 56.1 & 58.4 & 55.4 & 66.3 & 60.9 & 85.1 & 69.9 & 66.0 & 71.8 \\
\quad\quad w/ Page + OCR & 63.3 & 62.2 & 57.4 & 67.0 & 58.2 & 70.6 & 65.0 & \textbf{89.4} & \underline{76.9} & 68.5 & 76.5 \\
\quad\quad w/ Evidence & \underline{64.2} & 67.2 & 61.5 & 62.9 & 61.2 & 66.7 & 64.9 & \underline{89.1} & 76.4 & \underline{73.7} & \underline{78.2} \\
\quad\quad w/ Evidence + OCR & \textbf{67.4} & \textbf{68.8} & 63.1 & 64.3 & \textbf{63.6} & 66.8 & \underline{66.4$^{*}$} & 88.7 & 76.2 & \textbf{75.3} & \textbf{78.6} \\
\bottomrule
\end{tabular}

}
\caption{Main Results on MMLongBench-Doc and FinRAGBench-V. We report the accuracy of five types of evidence sources including pure text (TXT), layout (LAY), chart (CHA), table (TAB), and figure (FIG), and on unanswerable (UNA) samples. \textbf{Bold} and \underline{underlined} indicate the best and the second score per column. \dag{} denotes results reported in original papers, hence some results are unavailable. $^{*}$ Denotes results surpassing human experts (their performance on MMLongBench-Doc is 65.8).}
\label{tab:exp_main_results}
\end{table*}

\begin{table*}[tbp]
\centering
\normalsize
\setstretch{1.2}
\resizebox{\textwidth}{!}{
\begin{tabular}{lccccc|ccccc|c}
\toprule
\multirow{2}{*}{\textbf{Model}} 
& \multicolumn{5}{c|}{\textbf{FinRAGBench-V (subset w. bbox)}} 
& \multicolumn{5}{c|}{\textbf{MVToolbench (w.o. rotation)}} 
& \textbf{FinRAG (Neg)} \\
\cmidrule(lr){2-6} \cmidrule(lr){7-11}
& $\mathbf{Page_{acc}}$ & $\mathbf{Recall_{min}}$ & $\mathbf{Prec_{min}}$ & $\mathbf{F1_{min}}$ & $\mathbf{Recall_{EM}}$
& $\mathbf{Page_{acc}}$ & $\mathbf{Recall_{min}}$ & $\mathbf{Prec_{min}}$ & $\mathbf{F1_{min}}$ & $\mathbf{Recall_{EM}}$
& $\mathbf{Page_{acc}}$ \\
\midrule
Qwen3-VL-235B-A22B & 86.4 & 75.6 & 79.8 & 76.5 & 43.4 & 91.3 & 57.0 & 56.7 & 56.8 & 7.0 & 92.9 \\
Gemini-3-Pro-Preview & 94.6 & 83.1 & 82.2 & 81.3 & 41.3 & 98.6 & 70.3 & 67.8 & 68.6 & \textbf{31.8} & \textbf{96.8} \\
AgenticOCR-4B & \textbf{97.3} & \textbf{83.7} & 83.8 & 81.8 & \textbf{50.1} & 99.1 & \textbf{78.3} & 74.6 & \textbf{75.9} & 27.7 & 90.0 \\
AgenticOCR-8B & 96.8 & 83.3 & \textbf{85.6} & \textbf{82.8} & 44.0 & \textbf{99.4} & 75.9 & \textbf{74.8} & 75.2 & 25.5 & 93.1 \\
\bottomrule
\end{tabular}
}
\caption{Performance of AgenticOCR on FinRAGBench-V (subset with bounding boxes), MVToolbench (without rotation), and hard negative samples constructed from FinRAGBench-V. We report page-level accuracy and element-level metrics for positive samples, and page-level accuracy for negative samples. }
\label{tab:retrieval_results}
\end{table*}

\paragraph{Overall Performance.}
On MMLongBench-Doc, our 8B model with \textit{Evidence+OCR} input attains an overall accuracy of 66.4, which not only surpasses the human expert baseline of 65.8 (after adjusting for dataset size, our score corresponds to 65.9 on the original 1,082 samples) but also rivals the highly optimized DocLens with Gemini-2.5-Pro (67.6-1.5=66.1, subtracting about 1.5-point gain from adjudicator module). This demonstrates that \methodname{}, despite using a smaller agent backbone (Qwen3-VL-8B), can yield state-of-the-art results in long-form document understanding. On FinRAGBench-V, the same configuration achieves 78.6 accuracy, outperforming all prior agentic frameworks. This highlights the effectiveness of our query-driven extraction paradigm in dense, visually complex financial documents.

\paragraph{Strengths on Text, Layout, and Figure Modalities.}
\methodname{} exhibits particularly strong performance on text-only (TXT), layout (LAY), and figure (FIG) subsets of MMLongBench-Doc. Our 8B model with \textit{Evidence+OCR} reaches 67.4 on TXT, 68.8 on LAY, and 63.6 on FIG—surpassing the best DocLens configurations while using a significantly smaller model. This suggests that the agent's ability to precisely localize and extract relevant text blocks, preserve layout structure, and interpret figures via on-demand zoom is helpful for these modalities. The \textit{Evidence+OCR} input format, which interleaves low-resolution page snapshots, cropped visual patches and extracted text, provides a rich and compact representation that benefits both layout-sensitive and content-heavy reasoning.

\paragraph{Limitations on Tables and Unanswerable Questions.}
Despite overall gains, the current version of \methodname{} lags behind top-performing agentic frameworks on the table (TAB) and unanswerable (UNA) subsets. For tables, our model often returns incomplete extractions—e.g., only a single row or cell without the necessary headers or surrounding context—leading to incomplete evidence. Moreover, due to the lack of supervision on model-generated comments during training, the comments from AgenticOCR models occasionally introduce hallucinations that degrade answer quality, making the improvement of \textit{Evidence+OCR} marginal compared to \textit{Evidence}. On unanswerable questions, our model's accuracy (66.8) falls below that of DocLens (up to 74.0). The primary reason is the lower retrieval precision: DocLens retrieves on average only 1.5 pages per query, whereas our system retrieves 8.6 pages on average. The abundance of distracting pages, many of which contain superficially similar content, makes it challenging for the generator to confidently decide that no relevant evidence exists. This underscores not only the need for more quality control on comments content, retrieval precision and granularity, but also plenty of room for performance improvement.

\subsection{Ablation Studies and Discussions}
\label{sec:exp_diss}

\subsubsection{Performance Analysis of AgenticOCR Models}

\begin{figure*}[tb]
\centering
\includegraphics[width=\textwidth]{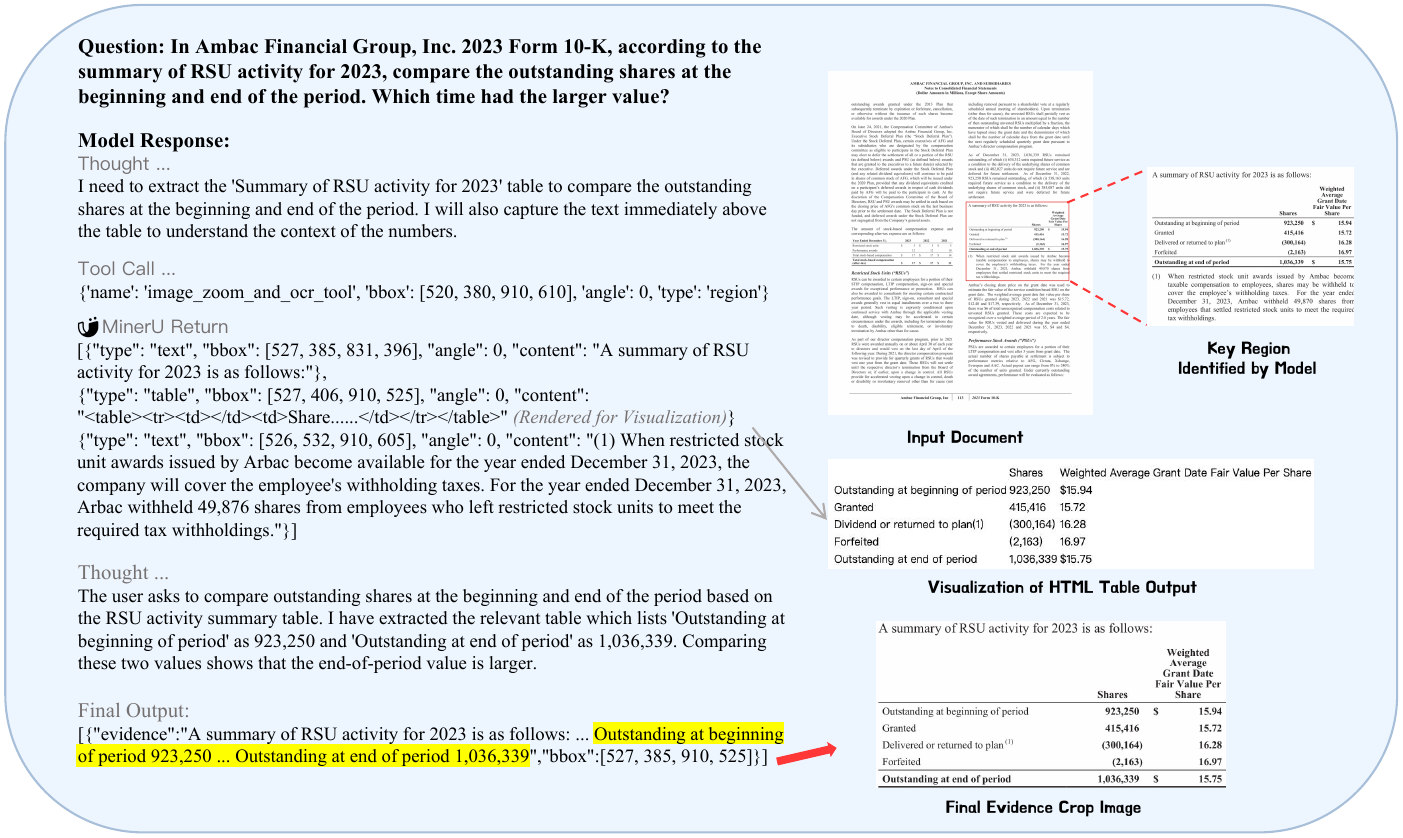}
\caption{
An example of the AgenticOCR Model retrieval and evidence extraction workflow. The figure illustrates the input document, the key region identified by the model, the visualization of structured HTML table output, and the final evidence crop image, demonstrating the system's ability to perform on-demand decompression of visual information through zoom and OCR operations.
}
\label{fig:retrieval_case_1}
\end{figure*}

To isolate \methodname{}'s evidence extraction capability demonstrated in Fig. \ref{fig:retrieval_case_1}, we conduct a dedicated evaluation on single-page scenarios. This assesses the model's ability to locate and discriminate relevant regions within a page, independent of the full RAG pipeline.

\textbf{Benchmarks.} We evaluate both page‑level relevance judgment and element‑level localization. For positive samples, we use two datasets with ground‑truth bounding boxes: (i) a subset of FinRAGBench‑V containing approximately 200 samples with human‑annotated boxes, and (ii) MVToolbench \citep{guo2025thinking}, which provides over 400 document pages paired with a query and a ground‑truth bounding box. To maintain a controlled setting, we revert any page rotations or flips present in the original MVToolbench, testing only on standard‑orientation pages. For negative samples, we construct 200 hard negatives from FinRAGBench‑V following the strategy described in Section~\ref{subsec:sft}: pages that are semantically or structurally similar to the query but contain no actual evidence.

\textbf{Metrics.} For positive samples, we compute:
\begin{itemize}
    \item \textbf{Page‑level accuracy}: \(\displaystyle \mathrm{Page}_{\mathrm{acc}} = \frac{\#\text{pages correctly judged as relevant}}{\#\text{total pages}}\).
    \item \textbf{Element‑level metrics}: \(\mathrm{Recall}_{\min}\) and \(\mathrm{Recall}_{\mathrm{EM}}\) as defined in Section~\ref{subsec:sft}. We also introduce \(\mathrm{Precision}_{\min}\) and \(\mathrm{F1}_{\min}\) to quantify over‑prediction:
    \begin{align}
        \mathrm{Precision}_{\min} &= \frac{1}{|\mathcal{B}^{\text{pred}}|} \sum_{b^{\text{pred}} \in \mathcal{B}^{\text{pred}}} \mathbb{I}\Big( \max_{b^{\text{gt}} \in \mathcal{B}^{\text{gt}}} \mathrm{IoU}_{\min}(b^{\text{gt}}, b^{\text{pred}}) \geq thres_{\min} \Big), \\
        \mathrm{F1}_{\min} &= \frac{2 \cdot \mathrm{Precision}_{\min} \cdot \mathrm{Recall}_{\min}}{\mathrm{Precision}_{\min} + \mathrm{Recall}_{\min}}.
    \end{align}
\end{itemize}
For negatives, we simply report page‑level accuracy as the fraction correctly classified as irrelevant.

\textbf{Baselines.} We compare \methodname-4B and \methodname-8B against strong models: Gemini-3-Pro-Preview (the teacher used for trajectory distillation) and Qwen3-VL-235B-A22B-Instruct. All models are tested with identical prompts, tool‑use protocols, and image resolutions. To account for the inherent randomness of LLM inference with temperature 1.0, we run each experiment 8 times and report the averaged results. The base Qwen3-VL-4B/8B-Instruct models are omitted because they exhibited poor instruction‑following and output‑format compliance, which would unfairly bias the comparison.

\textbf{Results.} Table~\ref{tab:retrieval_results} summarizes the performance. On positive samples, \methodname{} demonstrates strong capabilities in both page‑level relevance judgment and element‑level evidence extraction. For instance, \methodname-4B achieves a \(\mathrm{Page}_{\mathrm{acc}}\) of 97.3 and a \(\mathrm{Recall}_{\min}\) of 83.7 on the FinRAGBench‑V subset, indicating robust identification of relevant pages and retrieval of key evidence regions. Notably, it attains a \(\mathrm{Recall}_{\mathrm{EM}}\) of 50.1, approaching the quality of human‑annotated bounding boxes and suggesting that the model can produce reasonably precise matches under the strict exact‑match criterion. Overall, \methodname{} reaches a performance level comparable to Gemini-3-Pro-Preview. However, \methodname{} still exhibits a noticeable gap on hard negative samples, implying room for improvement in discriminating superficially similar yet irrelevant pages. Performance on extremely fine‑grained bounding box matching also remains less stable; for example, the relatively lower \(\mathrm{Recall}_{\mathrm{EM}}\) on MVToolbench indicates that precise box‑level alignment under exact‑match criteria continues to be challenging.

\begin{table}[tb]
\centering
\renewcommand{\arraystretch}{1.2}
\resizebox{\textwidth}{!}{%
\begin{tabular}{llcccc}
\toprule
\textbf{Dataset} & \textbf{Generator} & \textbf{Setting} & \textbf{Accuracy} & \textbf{\# Input Token} & \textbf{\# Output Token} \\
\midrule
\multirow{8}{*}{\begin{tabular}[c]{@{}l@{}}MMLongBench-Doc \\\\ Page-Level Metrics: \\ (Recall: 93.5, Precision: 28.0)\end{tabular}} & \multirow{4}{*}{Gemini-2.5-Pro} & Page & 60.9 & 3377.313 & 1403.871 \\
 &  & Page+OCR & 65.0 & 7613.253 & 1343.515 \\
 &  & Evidence & 64.9 & 7904.886 & 1399.083 \\
 &  & Evidence+OCR & 66.4 & 9476.468 & 1409.607 \\
\cmidrule{2-6}
 & \multirow{4}{*}{Qwen3-VL-32B-Thinking} & Page & 62.1 & 9938.155 & 1326.319 \\
 &  & Page+OCR & 62.6 & 14517.723 & 1300.137 \\
 &  & Evidence & 63.6 & 12649.456 & 1371.042 \\
 &  & Evidence+OCR & 63.4 & 13238.009 & 1344.278 \\
\midrule
\multirow{8}{*}{\begin{tabular}[c]{@{}l@{}}FinRAGBench-V \\\\ Page-Level Metrics: \\ (Recall: 95.3, Precision: 34.8)\end{tabular}} & \multirow{4}{*}{Gemini-2.5-Pro} & Page & 71.8 & 3042.460 & 1930.657 \\
 &  & Page+OCR & 76.5 & 9289.288 & 1721.593 \\
 &  & Evidence & 78.2 & 6474.760 & 1849.603 \\
 &  & Evidence+OCR & 78.6 & 9090.601 & 1832.599 \\
\cmidrule{2-6}
 & \multirow{4}{*}{Qwen3-VL-32B-Thinking} & Page & 73.9 & 8353.006 & 1755.662 \\
 &  & Page+OCR & 75.7 & 15465.149 & 1595.539 \\
 &  & Evidence & 78.2 & 11189.823 & 1622.726 \\
 &  & Evidence+OCR & 77.6 & 14076.330 & 1688.552 \\
\bottomrule
\end{tabular}
}
\caption{Performance of the visual retrieval pipeline on MMLongBench-Doc and FinRAGBench-V datasets. We report accuracy, average input tokens, and output tokens for different input configurations (Page, Page+OCR, Evidence, Evidence+OCR) using two generators: Gemini-2.5-Pro and Qwen3-VL-32B-Thinking. For Gemini, the \textit{Evidence} and \textit{Evidence+OCR} settings incur high token costs due to its fixed per-image token allocation policy, whereas Qwen3-VL allows finer control, demonstrating the efficiency potential of AgenticOCR.}
\label{tab:visual_retrieval}
\end{table}

\subsubsection{Performance Analysis of Visual Retrieval Pipeline}

To understand the system-level efficiency of our framework, we analyze the performance of the visual retrieval pipeline alongside the token consumption of downstream generators, as detailed in Table~\ref{tab:visual_retrieval}.

First, the results demonstrate that \methodname{} has the potential to achieve higher final generation accuracy while operating at a lower generative token consumption compared to full-page processing baselines. For example, on FinRAGBench-V benchmark and Gemini-2.5-Pro generator, \textit{Evidence+OCR} reduces token consumption from \textit{Page+OCR}'s 9,289 into 9,090, and improves accuracy from 76.5 into 78.6. By selectively cropping relevant regions and providing on-demand text extraction, our pipeline filters out irrelevant visual noise and maximizes the signal-to-token ratio. 

However, an apparent anomaly is observed when using Gemini-2.5-Pro on MMLongBench-Doc: the token consumption for the \textit{Evidence} and \textit{Evidence+OCR} settings is surprisingly higher than that of the \textit{Page+OCR} baseline. This counter-intuitive token inflation stems from the inherent image token allocation mechanism of the Gemini-2.5-Pro models. Currently, the Gemini 2 series employs specialized Pan \& Scan pre-processing for input images\footnote{See \url{https://ai.google.dev/gemini-api/docs/media-resolution} for details on Gemini's media resolution handling.}, and it does not allow users to independently specify or constrain the visual token budget for individual image crops. Consequently, feeding multiple cropped evidence patches accumulates a large visual token footprint. To validate that \methodname{} inherently reduces context redundancy, we evaluate Qwen3-VL-32B-Thinking as an alternative generator. As shown in Table~\ref{tab:visual_retrieval} on MMLongBench-Doc, utilizing \textit{Evidence+OCR} with Qwen3-VL effectively reduces the total input tokens compared to the \textit{Page+OCR} setting (from 14,517 down to 13,238 tokens), validating the token-saving potential of our agentic extraction approach.

Finally, we analyze the retrieval quality. While the page-level recall is high (93.5\% on MMLongBench-Doc and 95.3\% on FinRAGBench-V), the precision remains relatively low (28.0\% and 34.8\%, respectively). This indicates that a significant number of irrelevant pages are still passed to the downstream generator. However, it is worth noting that despite this noise, the overall token volume remains around the 10,000-token magnitude. Because this length fits comfortably within the effective long-context processing windows of both Gemini-2.5-Pro and Qwen3-VL-32B-Thinking, the models can successfully filter out distracting information. As a result, the limited retrieval precision does not severely degrade the final answer accuracy, though improving precision remains a promising direction.

\section{Conclusion}
\label{sec:conclusion}

In this work, we introduced \methodname{}, a paradigm shift that transitions optical character recognition from a static, full-page preprocessing step into a dynamic, query-driven agentic process. By equipping VLMs with specialized visual interaction tools, \methodname{} autonomously localizes, crops, and parses only the query-relevant regions within complex visual documents. Our two-stage training pipeline—combining supervised fine-tuning via trajectory distillation with GRPO-based reinforcement learning—yields robust models capable of performing on-demand decompression of visual information. Extensive experiments on challenging benchmarks, including MMLongBench-Doc and FinRAGBench-V, demonstrate that integrating \methodname{} as a plug-and-play middleware can minimize visual noise, maximize the signal-to-token ratio, and ultimately boost both the accuracy and efficiency of downstream generation in visual RAG systems. 

Despite these promising results, we position \methodname{} to be a preliminary exploration and several limitations remain to be addressed. We hope this work serves as a stepping stone to inspire the community to further explore the agentic parsing paradigm. Future directions could include building structured index for page retrieval, enhancing retrieval precision via scaling up data engineering, and exploring tighter integrations between visual agents and generative LLMs.


\newpage 


\clearpage
\newpage
{ 
\bibliographystyle{plainnat}
\setcitestyle{numbers}
\bibliography{paper}
}

\clearpage
\newpage
\beginappendix

\section{Prompt Templates}
In this section, we provide the prompts in our framework.
\label{sec:app_prompt}

\begin{promptbox}[breakable]{Prompt for Qwen3-VL-Reranker-8B}
Given a search query, retrieve relevant candidates that answer the query. Note that 'Page ID' indicates the physical page index in the document file, which does not necessarily correspond to the logical page number printed on the page image.
\end{promptbox}

\begin{promptbox}[breakable]{Prompt for AgenticOCR models}
You are an advanced Visual Document Analysis Agent capable of precise evidence extraction from document images. Your goal is to answer user queries by locating, reading, and extracting specific information from a page.

### Your Capabilities & Tools

You have access to a powerful tool named **`image_zoom_and_ocr_tool`**.

* **Functionality**:
  Crop a specific region of the image, optionally rotate it, and perform OCR or layout analysis depending on the element type.

* **When to use**:

  * Always use this tool when the user asks for **specific text, numbers, names, dates, tables, equations, images, or factual details** from the page.
  * If the target text or table is rotated, estimate and set the `angle` parameter before cropping.
  * Use `type` to indicate the content type for the region:

    * `"region"` — perform layout detection + OCR on a potentially complex area, returning detailed structured information.
    * `"text"` — perform OCR for a single text element.
    * `"table"` — perform OCR and parsing for a table element.
    * `"image"` — crop the image region only, without OCR.
    * `"equation"` — perform OCR for mathematical or scientific equations.

* **Parameters**:

  * `label`: A short description of what you are looking for.
  * `bbox`: `[xmin, ymin, xmax, ymax]` in **0–1000 normalized coordinates**, relative to the original page.
  * `angle`: Rotation angle (counter-clockwise) applied after cropping. Always try to adjust it so the content is upright for best recognition.
  * `type`: One of `"region"`, `"text"`, `"table"`, `"image"`, `"equation"`.

### Tool Usage Example

Use the tool strictly in the following format:
<think>
...
</think>
<tool_call> 
{"name": "image_zoom_and_ocr_tool", "arguments": {"label": "<A short description of what you are looking for>", "bbox": [xmin, ymin, xmax, ymax], "angle":<0/90/180/270>, "type": "<region/text/table/image/equation>"}}
</tool_call>

### Your Input and Task

The input includes:

1. One page image of a visual document.
2. The user's query intent.

Please execute the following steps:

1. **Semantic Matching**: Carefully observe the image to determine if the page content contains evidence information relevant to the user's query. If it is irrelevant, return an empty list.
2. **Precise Localization**: If relevant, extract the complete chain of visual evidence that helps to answer the query (text blocks, tables, charts, images, or equations).
3. **Special Notes**: The page image may contain several evidence pieces. Pay attention to tables, charts, images, and equations, as they could also contain evidence.

### Output Format

After gathering information, output the list of relevant evidence in the following JSON format.
<think>
...
</think>
```json
[
  {
    "evidence": "<self-contained content, understandable without page context>",
    "bbox": [xmin, ymin, xmax, ymax] # 0-1000 normalized coordinates 
  }
  ...
]
```

If the page image is not relevant, return an empty list.
<think>
...
</think>
```json
[]
```
\end{promptbox}

\newpage
\begin{promptbox}[breakable]{Prompt for the Gemini-2.5-Pro generator}
## ROLE
You are an expert AI assistant specializing in multimodal long document understanding. Your task is to carefully analyze the provided page images (which may contain text, figures, tables, and other content) and provide a precise answer to the user's question.
## Follow these instructions carefully:
- Core Objective: Your primary goal is to accurately and concisely answer the user's question based on the content of the provided document pages.
- Rules of numerical answers:
    - If the user asks for an absolute number (e.g., with questions like "How many...?"), you must first attempt to locate the number directly. If it cannot be found, find the relevant percentage and total count (or other necessary data) to calculate the absolute number. If the calculated absolute number for discrete entities (e.g., people, companies, objects) is a decimal, you must round it to the nearest whole number.
    - If the user asks for a percentage (or proportion), you must first attempt to locate the percentage directly. If it cannot be found, find the absolute numbers of the subgroup and the total count (or other necessary data) to calculate the percentage.
    - If the user's question is ambiguous and does not explicitly specify a number or percentage (e.g., "What's the gap between...?"), you must default to providing the absolute value. If you can only find relative values (percentages) in the chart, you must make every effort to find a total number within the provided context to calculate the absolute value. Only return the relative value as a last resort if a total number cannot be found, and explain that you cannot find total number in this case.
- Zoom-in Feature: When a page image contains figures or tables and requires closer inspection, we may provide zoomed-in images of these elements, appended after the main page image, to help you examine them closely. We will also extract text from the page image into Markdown format. Note: For questions related to page layout, you must refer to the original page image itself, not the zoomed-in images or the Markdown text, as they may lose layout information. For instance, if asked for the first figure on the page, you should consult the full page image to determine its order, not the sequence of the provided zoomed-in images.
- Page Numbering: Page numbers in the user's question typically refer to the number printed on the page image, not the page's index in the document file. For example, if a PDF's first page is the cover and the third page is the first page of content (labeled "Page 1"), a user's question about "page 1" refers to that third page. Similarly, when asked to provide a page number, you should return the printed page number from the image. Only return the page index if no number is printed on the page.
- Rule of faithfulness: Be faithful. If the provided pages do not contain sufficient information to answer the user's question, you should answer `Not answerable`. For example, if the user asks for a man in green shirts, but there are only man in red shirts in the provided pages, you should answer `Not answerable`; if the user asks for the boy playing badminton, but there are only boys playing football in the provided pages, you should answer `Not answerable`; if the user asks for a certain year's data but the provided pages only contain data for other years, you should answer `Not answerable`; if the user asks for the color of a certain object but the provided pages do not contain that object, you should answer `Not answerable`. 

### Input Format
The user will provide:
- A set of evidence (images and text) retrieved from the documents.
- The question to answer.

## Output Format:
Your entire response MUST be a single, valid JSON object and nothing else. Do not wrap it in markdown code blocks or add any other text. The JSON object must contain exactly two fields: analysis (string), and prediction (string).
- analysis field: Briefly explain your thought process. Describe how you located the answer within the document, which pages, tables, or figures you referenced, and how you connected the information to the question.
- prediction field: This must be a string containing the direct answer to the user's question.
\end{promptbox}

\section{Dataset Statistics}
\label{sec:app_data}

Statistics of MMLongBench-Doc and FinRAGBench-V is presented in Table~\ref{tab:data_statistics}. MMLongBench-Doc comprises documents from 7 various domains, including Research report/Introduction, Tutorial/Workshop, Academic Paper, Guidebook, Brochure, Administration/Industry file, Financial Report. FinRAGBench-V, on the other hand, focuses sole on financial reports. Note that, 8 mismatched PDF-question pairs from MMLongBench-Doc are removed in our experiments.

\begin{table}[h!]
\centering
\footnotesize
\renewcommand{\arraystretch}{1.1}
\setlength\tabcolsep{14pt}
\begin{tabular}{l|cc}
\toprule
\multirow{2.5}{*}{\textbf{Statistics}}  &  \multicolumn{2}{c}{\textbf{Dataset Name}} \\  \cmidrule(l){2-3}
& MMLongBench-Doc & FinRAGBench-V \\ \midrule
\textbf{Documents} & 135 & 301 \\
- Average/Medium pages & 47.5 / 28 & 76.1 / 57.0 \\
- Average/Medium words & 8,393 / 5,743 & 36,026 / 16,329 \\ \midrule
\textbf{Total question} & 1,082 (-8=1,074) & 1,394 \\
- Single-page question & 494 (45.7\%) & 1,218 (87.4\%) \\
- Cross-page question & 365 (33.7\%) & 178 (12.6\%) \\
- Unanswerable question & 223 (20.6\%) &  - \\ \midrule
\textbf{Evidence source} & & \\
- Pure-text & 305 (35.5\%) & 302 (21.7\%) \\
- Layout & 119 (13.9\%) & - \\
- Table & 218 (25.4\%) & 573 (41.1\%) \\
- Chart & 178 (20.7\%) & 519 (37.2\%) \\
- Image & 304 (35.4\%) & - \\ \midrule
Avg. / Max. question words & 16.2 / 54 & 35.8 / 108\\
Avg. / Max. answer words & 2.1 / 66 & 23.4 / 174 \\
\bottomrule
\end{tabular}
\caption{Statistics of benchmarks}
\label{tab:data_statistics}
\end{table}

\section{Implementation Details}
\label{sec:app_implement}

\subsection{Evaluation on MMLongBench-Doc}
\label{sec:app_mmlong_eval}
For MMLongBench-Doc, we fixed some annotations about evidence pages, following DocLens \citep{zhu2025doclens}. We thank the authors for providing corrected datasets, helpful discussions and other artifacts \citep{zhu2026paperbanana}. Note that, 8 mismatched PDF-question pairs from MMLongBench-Doc are further removed in our experiments (PDF File: dr-vorapptchapter1emissionsources-121120210508-phpapp02\_95.pdf), resulting into 1,074 test pairs. 

\begin{promptbox}[breakable]{Prompt used for Answer Extraction on MMLongBench-Doc}
Given the question and analysis, you are tasked to extract answers with required formats from the free-form analysis. 
- Your extracted answers should be one of the following formats: (1) Integer, (2) Float, (3) String and (4) List. If you find the analysis and the question can not be answered from the given documents, type "Not answerable". Exception: If the analysis only tells you that it can not read/understand the images or documents, type "Fail to answer".
- Please make your response as concise as possible. Also note that your response should be formatted as below:
```
Extracted answer: [answer]
Answer format: [answer format]
```

Please read the following example, then extract the answer from the model response and type it at the end of the prompt. 

---
Question: List the primary questions asked about the services in this report.
Analysis:  The primary questions asked about the services in the report for The Limes Residential Home are:\n\n1. Is the service safe?\n2. Is the service effective?\n3. Is the service caring?\n4. Is the service responsive?\n5. Is the service well-led?
Extracted answer: ['Is the servife safe?', 'Is the service effective', 'Is the serve caring?', 'Is the service responsive?', 'Is the service well-led?']
Answer format: List

---
Question: How many regulations of the HSCA 2008 are breached in all according to this report?
Analysis: According to the report, the provider breached 10 Health and Social Care Act 2008 (Regulated Activities) Regulations in total. Here are the specifics:\n\n1. Regulation 13: Safeguarding service users from abuse and improper treatment\n2. Regulation 12: Safe care and treatment\n3. Regulation 18: Staffing\n4. Regulation 11: Need for consent\n5. Regulation 10: Dignity and respect\n6. Regulation 9: Person-centred care\n7. Regulation 17: Good governance\n8. Regulation 18 (CQC Registration Regulations 2009): Notification of other incidents\n9. Regulation 18: Failure to maintain an accurate and up-to-date care plan\n10. Regulation 11: Failure to implement the Mental Capacity Act 2005 code of practice effectively\n\nThese breaches involve issues concerning staffing, safeguarding, medicines management, dignity and respect, consent, care planning, governance, and failure to notify the CQC of incidents.
Extracted answer: 10
Answer format: Integer

---
Question: According to the survey that is the percentage of Chinese who are paying more or about the same attention to politics after Trump's election?
Analysis: The survey provided does not specify the percentage of Chinese individuals specifically who are paying more or about the same attention to politics after Trump's election. The report focuses primarily on American demographics and does not include specific details about the Chinese population in relation to this question. If you need information about a different demographic or a summary of the findings from the American demographic, I can certainly help with that!
Extracted answer: Not answerable
Answer format: String

---
Question: How many quotations from male respondent over 50 years old are included in this report?
Analysis: The image you've provided appears to be a screenshot of a document with multiple charts. However, the text is too small and blurry to read accurately. If you can provide a clearer image or more context, I might be able to help you with your question.
Extracted answer: Fail to answer
Answer format: String

---
\end{promptbox}

\subsection{Evaluation on FinRAGBench-V}
\label{sec:app_finrag_eval}
The final score is 1.0 if the predicted SCORE is at least 4, otherwise 0.
\begin{promptbox}[unbreakable]{Prompt used for Answer Evaluation on FinRAGBench-V}
You are an expert evaluation system for a question answering chatbot.

You are given the following information:
- a user query and reference answer
- a generated answer

Your job is to judge the relevance and correctness of the generated answer.
Output a single score that represents a holistic evaluation.
You must return your response in a line with only the score.
Do not return answers in any other format.
On a separate line provide your reasoning before the score as well.

Follow these guidelines for scoring:
- Your score has to be between 1 and 5, where 1 is the worst and 5 is the best.
- If the generated answer is not relevant to the user query, you should give a score of 1.
- If the generated answer is relevant but contains factual errors or significant hallucinations, you should give a score between 2 and 3.
- If the generated answer is relevant and fully correct according to the reference, you should give a score between 4 and 5.

**Special Instruction for Open-Ended Questions:**
- If the user query is open-ended (allowing for multiple valid answers), the generated answer DOES NOT need to match the reference answer exactly.
- As long as the generated answer is highly relevant and contains no factual conflicts with the reference, give a high score.

**Special Instruction for Numerical and Logical Equivalence:**
- If the generated answer represents the **same value or concept** as the reference answer but in a different format, unit, or perspective, it MUST be considered CORRECT.
- **Unit Conversion:** (e.g., Reference: "1 kilometer", Generated: "1000 meters" -> Correct).
- **Format Differences:** (e.g., Reference: "0.5", Generated: "50%" or "1/2" -> Correct).
- **Absolute vs. Relative:** If the generated answer uses a relative value (e.g., percentage) while the reference uses an absolute value (or vice versa), and they are mathematically consistent based on the context, treat it as correct.
- You should perform necessary mental calculations to verify if the generated answer can be derived from the reference or the reference can be derived from the generated answer.

Example Response:
REASON: The generated answer uses '500 meters' while the reference says '0.5 km'. These are mathematically equivalent, so the answer is correct.
SCORE: 5

User:
## User Query
{query_text}

## Reference Answer
{expected_answer}

## Generated Answer
{actual_answer}
\end{promptbox}

\section{Case Studies}

\begin{figure*}[h]
\centering
\includegraphics[width=\textwidth]{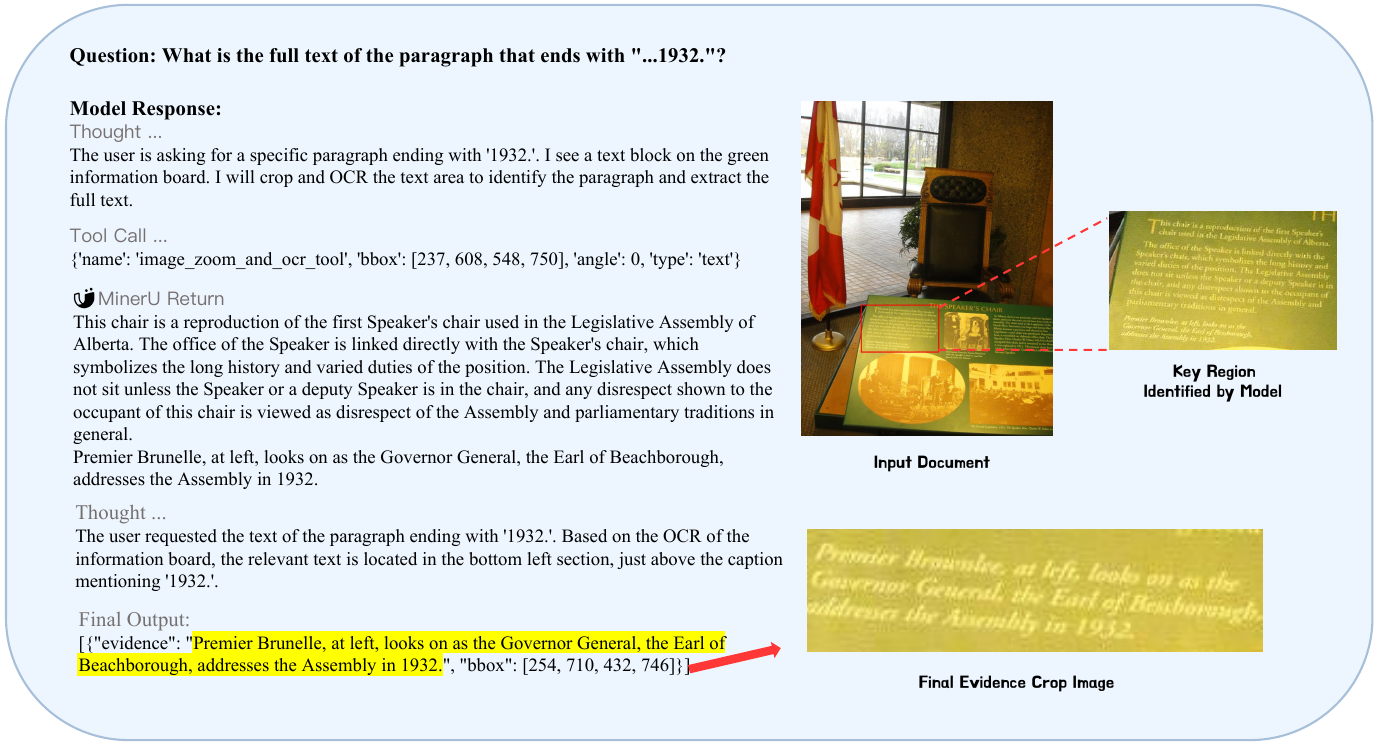}
\caption{
Another example of the AgenticOCR Model retrieval and evidence extraction workflow from MVToolbench. Compared with Figure \ref{fig:retrieval_case_1}, the AgenticOCR model invokes \texttt{image\_zoom\_and\_ocr\_tool} in "text" mode, which returns plain textual content. The figure presents the input document, the model-identified key region, and the final evidence crop corresponding to the extracted sentence. AgenticOCR successfully completes the task in a real-world scenario, demonstrating its capability to accurately localize and retrieve evidence through on-demand zoom and OCR operations.
}
\label{fig:retrieval_case_2}
\end{figure*}

\begin{figure*}[h]
\centering
\includegraphics[width=0.9\textwidth]{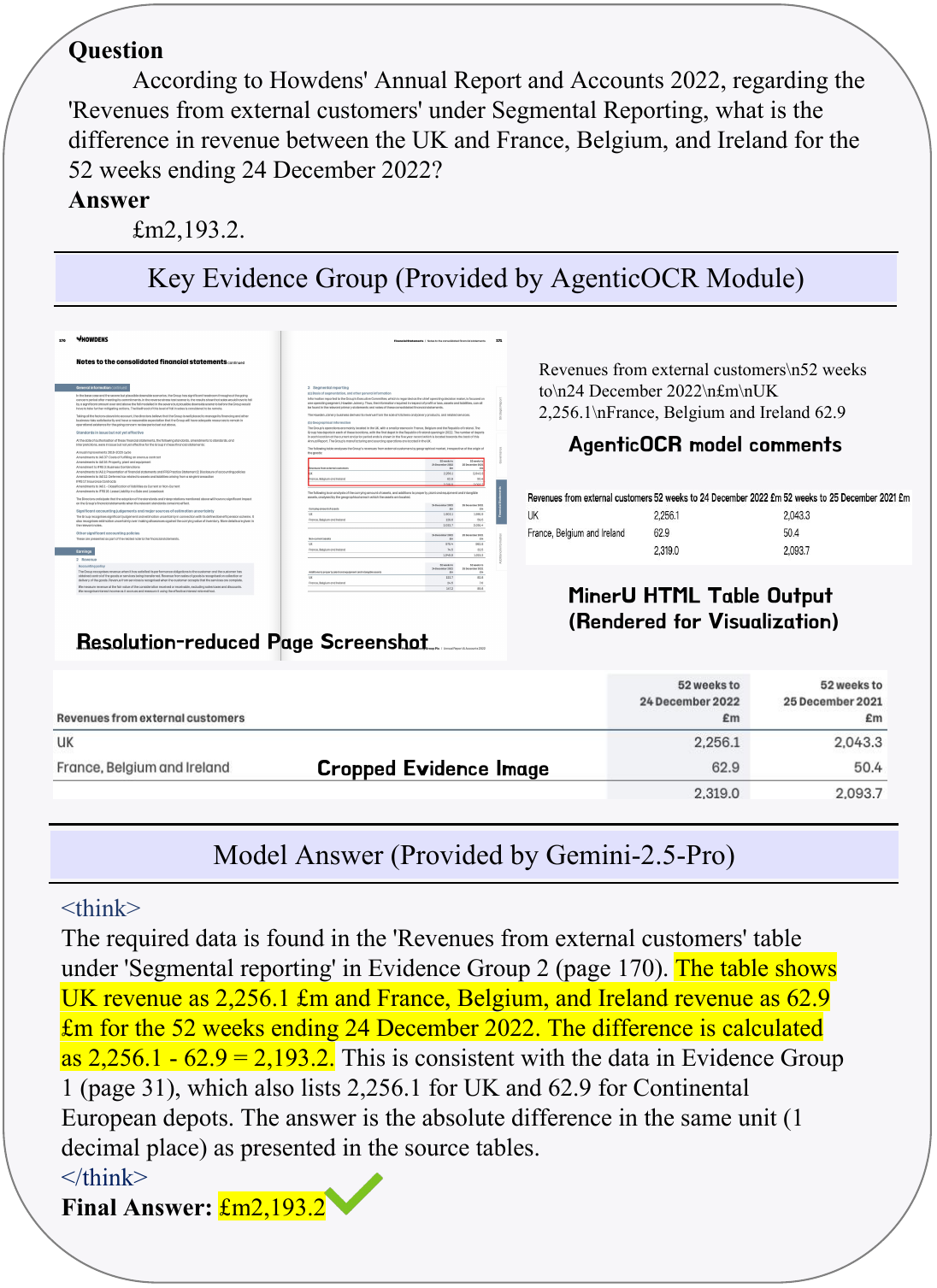}
\caption{
An example where a generative model (Gemini-2.5-Pro) answers a question based on the evidence groups provided by the AgenticOCR Module. The model derives the correct result through simple calculations using the key table information and even performs basic cross-validation across different evidence groups.
}
\label{fig:generate_case_1}
\end{figure*}

\begin{figure*}[h]
\centering
\includegraphics[width=0.9\textwidth]{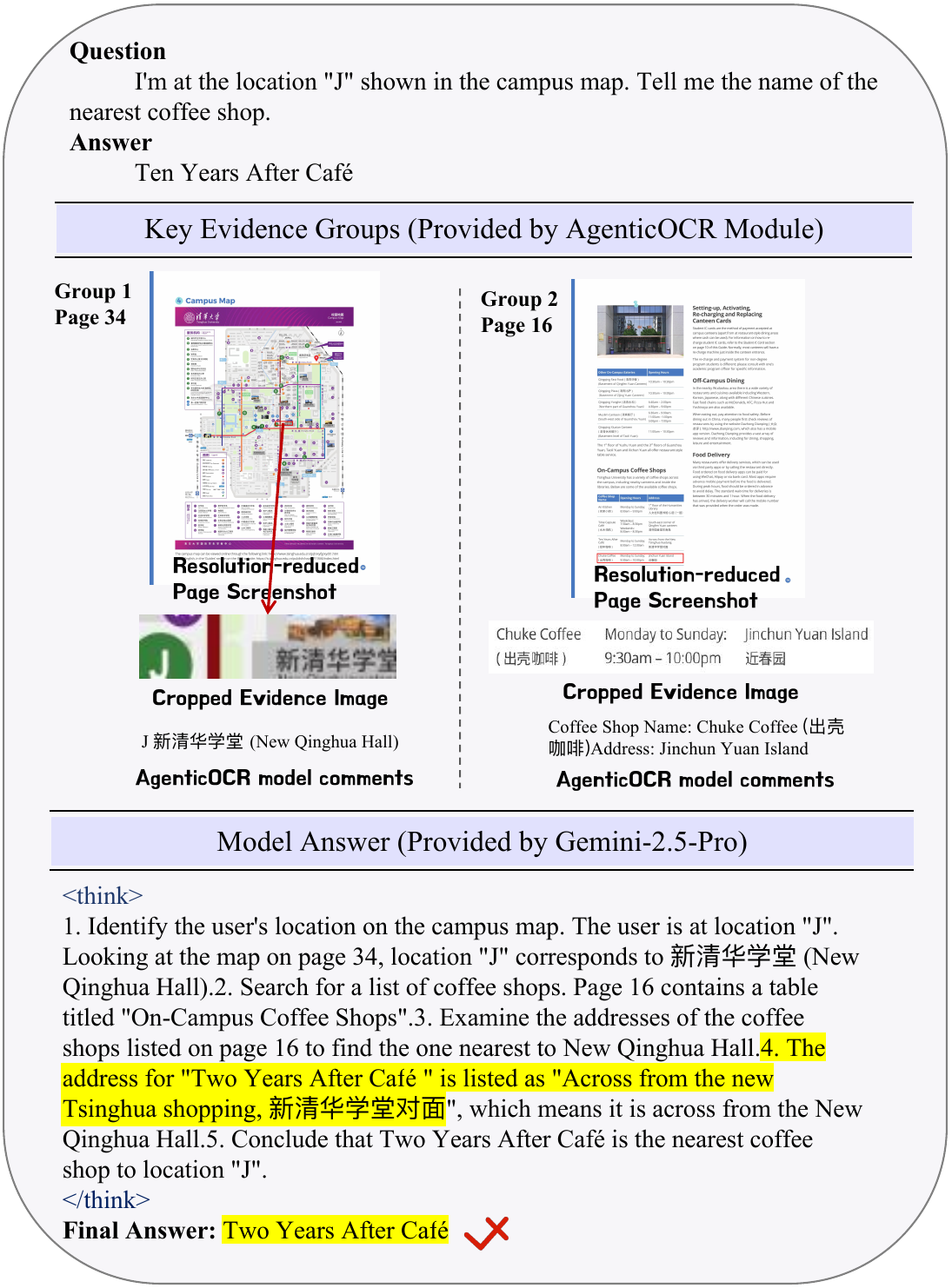}
\caption{
A hard case in our Module. Since AgenticOCR Model is limited to single-page processing, it selected an incorrect key region (the table in Group 2 / Page 16 contains two cafés). Although the generative model demonstrates relatively strong robustness to distractors, it misidentified the café name due to the low-resolution page screenshot, resulting in an incorrect final answer. This example highlights the current limitations of our framework and points to directions for future improvement.}
\label{fig:generate_case_2}
\end{figure*}

\end{document}